\pdfoutput=1

\documentclass[11pt]{article}

\usepackage{acl}

\usepackage{times}
\usepackage{latexsym}

\usepackage[T1]{fontenc}

\usepackage[utf8]{inputenc}

\usepackage{microtype}

\usepackage{booktabs}
\usepackage{amsmath,amssymb}
\usepackage{graphicx}
\usepackage{array}
\usepackage{enumitem}
\usepackage{color}
\usepackage{caption}
\usepackage{subcaption}
\usepackage{pifont}
\usepackage{multirow}
\newcolumntype{L}[1]{>{\raggedright\let\newline\\\arraybackslash\hspace{0pt}}m{#1}}
\newcolumntype{C}[1]{>{\centering\let\newline\\\arraybackslash\hspace{0pt}}m{#1}}
\newcolumntype{R}[1]{>{\raggedleft\let\newline\\\arraybackslash\hspace{0pt}}m{#1}}

\usepackage{pdfpages}
\usepackage{listings}

\definecolor{codegreen}{rgb}{0,0.6,0}
\definecolor{codegray}{rgb}{0.5,0.5,0.5}
\definecolor{codepurple}{rgb}{0.58,0,0.82}
\definecolor{backcolour}{rgb}{0.95,0.95,0.92}
\definecolor{OliveGreen}{rgb}{0.0039,0.4745,0.4353} 
\definecolor{Bittersweet}{rgb}{0.996,0.435,0.369}

\lstdefinestyle{mystyle}{
    backgroundcolor=\color{backcolour},   
    commentstyle=\color{codegreen},
    keywordstyle=\color{magenta},
    numberstyle=\tiny\color{codegray},
    stringstyle=\color{codepurple},
    basicstyle=\ttfamily\footnotesize,
    breakatwhitespace=false,         
    breaklines=true,                 
    captionpos=b,                    
    keepspaces=true,                 
    numbers=left,                    
    numbersep=5pt,                  
    showspaces=false,                
    showstringspaces=false,
    showtabs=false,                  
    tabsize=2
}

\title{Representation Learning for Conversational Data using Discourse Mutual Information Maximization}  
\author{
    Bishal Santra\textsuperscript{\ding{171}}\\
    IIT Kharagpur
    \And 
    Sumegh Roychowdhury\\
    IIT Kharagpur
    \And 
    Aishik Mandal\\
    IIT Kharagpur
    \AND 
    Vasu Gurram\\
    IIT Kharagpur
    \And
    Atharva Naik \\
    IIT Kharagpur
    \And
    Manish Gupta\textsuperscript{\ding{168}}\\
    Microsoft, India
    \And
    Pawan Goyal\textsuperscript{\ding{169}}\\
    IIT Kharagpur
    \AND
    {\normalfont \textsuperscript{\ding{171}}bsantraigi \ding{41} gmail.com, \ding{168} gmanish \ding{41} microsoft.com, \ding{169} pawang \ding{41} cse.iitkgp.ac.in}
}

\begin{document}
\maketitle

\begin{figure*}[!thb]
  \centering
  \includegraphics[width=\textwidth]{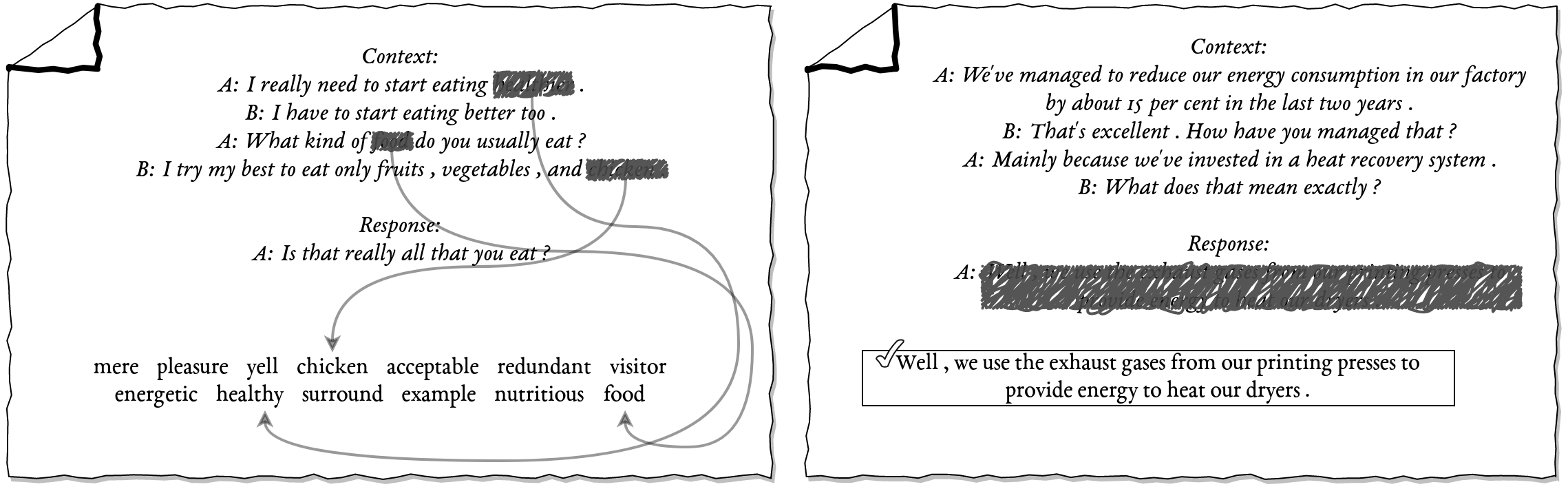}
  \caption{Possible reasoning involved in two types of pretraining: Word-level (left), Discourse-level (right). In a discourse-level reasoning task, the immediately preceding utterance may not be enough for understanding the full context. To predict the correct response, the model will need to capture both the larger context, in this case the topic of discussion, and the intent (e.g., asking for details) of the preceding utterance. In comparison, word-level reasoning is often easier and can be solved using local reasoning.
  Each of the three masked-words, in the left image, could have been predicted with reasonable confidence without any more information than the utterance itself.}
  \label{fig:main-idea}
\end{figure*}
\begin{abstract}
Although many pretrained models exist for text or images, there have been relatively fewer attempts to train representations specifically for dialog understanding. Prior works usually relied on finetuned representations based on generic text representation models like BERT or GPT-2. But such language modeling pretraining objectives do not take the structural information of conversational text into consideration. 
Although generative dialog models can learn structural features too, we argue that the structure-unaware word-by-word generation is not suitable for effective conversation modeling.
We empirically demonstrate that such representations do not perform consistently across various dialog understanding tasks. 
Hence, we propose a structure-aware Mutual Information based loss-function DMI (Discourse Mutual Information) for training dialog-representation models, that additionally captures the inherent uncertainty in response prediction.
Extensive evaluation on nine diverse dialog modeling tasks shows that our proposed DMI-based models outperform strong baselines by significant margins.
\end{abstract}

\section{Introduction}
Representation learning has transformed how we can apply machine learning to solve real-world problems. However, despite a vast body of research on pretrained language representations, there have been relatively fewer attempts to train representations specifically for dialog understanding. Prior works mostly relied on finetuned representations based on generic models like BERT~\cite{devlin2018bert} or GPT-2~\cite{radford2019language}. In our experiments, we demonstrate that such representations do not perform uniformly across various dialog understanding tasks such as dialog-act classification, intent detection or dialog evaluation. 

On the other hand, prior works on pretraining large-scale dialog models focused mainly on open-domain generation. These works evaluated their models only on dialog generation \cite{zhang2019dialogpt,roller2020recipes,adiwardana2020towards} or tasks related directly to the pretraining objective \cite{Henderson2020ConveRTEA,gao2020dialogrpt}. Their effectiveness on other dialog understanding tasks like act classification or intent detection remains unexplored. 
So we ask the following research question: \textit{Can we learn enriched representations directly at the pretraining phase that are specifically helpful for dialog understanding?}


Existing language modeling (causal or masked) pretraining objectives unfortunately are not the best to model dialogs for these reasons: (1) The model is not directly trained to learn the content discourse structure (e.g., context-response in dialogs). (2) Such models are trained to generate the response word-by-word rather than predicting a larger unit. (3) The inherent one-to-many nature of dialog generation implies that the encoding model should be able to capture uncertainty in the response prediction task, that such models ignore. 

Hence, in this paper, we propose pretraining objectives for improved dialog modeling that turn the discourse-level organizational structure of texts from natural sources (e.g., documents, dialogs, or monologues) into a learnable objective. We call this objective the Discourse Mutual Information (DMI). The key insight towards the design of our pretraining objective is to 
capture representations that can account for a meaningful conversation out of a specific ordered sequences of utterances. We hope that a discourse-level pretraining objective with conversational data would guide the model to learn complex context-level features. 
For example, in Fig.~\ref{fig:main-idea}, we illustrate the differences between standard language modeling (causal or masked) based pretraining objectives and a discourse-level reasoning task.

The second research question that we ask is \textit{whether discourse-level features learned using self-supervised pretraining outperform word-level pretraining objectives for downstream dialog understanding tasks.}
Experimentally, we show that representations learned using the proposed objective function are highly effective compared to both existing discriminative as well as generative dialog models.
In terms of various dialog understanding tasks, our models achieve state-of-the-art performances in several tasks (absolute improvements up to 8.5\% and 3.5\% in task accuracies in probing and finetuning setups, resp.) and perform consistently well across a variety of dialog understanding tasks, whereas baseline models usually have a rather imbalanced performance across tasks.

Overall, our main contributions are as follows. 
 \begin{itemize}
   \item We propose DMI, a novel information-theoretic objective function for pretraining dialog representation. 
  \item We release pretrained dialog-representation models in three different sizes (small, medium and base) based on our proposed self-supervised learning objectives\footnote{To access the pretrained dialog representation models and the source codes, please visit \url{https://bsantraigi.github.io/DMI}}. 
  \item We extensively evaluate our DMI based representations on multiple open-domain downstream tasks like intent detection, dialog-act classification, response retrieval, dialog reasoning, and response-generation evaluation, and beat state-of-the-art across nine tasks in both probe as well as finetune setups.
\end{itemize}

\section{Literature Review}
\label{sec:lit}

\begin{table*}[!thb]
\centering
\scriptsize
\begin{tabular}{@{}lR{3.3cm}R{2.56cm}R{1.76cm}R{1.cm}r@{}}
\toprule
\textbf{Model} & \textbf{Training Data Size} & \textbf{Pretraining Obj.} & \textbf{Architecture} & \textbf{Param} & \textbf{Downstream Task} \\ \midrule
\textit{DialoGPT\_Small} & 147M Dialogs & CE & GPT-2 & 125M & Generation /w MMI \\
\textit{DialogRPT} & 133M CR pairs & Response Ranking & DialoGPT & 345M & Human Feedback Prediction \\
\textit{Blenderbot\_Small} & 1.5B comments & CE & Tr. S2S & 90M & Generation \\
\textit{Meena $\ddag$} & 40B words, 341 GB text & CE & Evolved Tr. S2S & 2.4B & Generation \\
\textit{ContextPretrain $\ddag$} & 10k Dialogs, MultiWoz & NUR, NUG, MUR, I2 & HRED & - & Multiwoz (DST, Act, NUG, NUR) \\
\textit{DEB} & 727M Dialogs & MLM, NSP & BERT & 110M & Adv/Random Dialog Evaluation \\
\textit{ConveRT $\dagger$} & 727M Dialogs & Response Selection & Tr. Encoder & 29M & Response Selection \\
\textit{ConvFiT $\ddag$} & 8\% of 727M Dialogs + Intent data & Response Selection & BERT & 110M & Intent Detection \\
\textit{DMI\_Base} & 7.5-10\% of 727M Dialogs & InfoNCE-S & Tr. Encoder & 124M & 9 Dialog-NLU tasks \\ \bottomrule
\end{tabular}%
\caption{Survey of Pretrained Dialog Models. NUR: next utterance 
retrieval, NUG: next utterance generation, MUR: masked utterance retrieval, 
I2: inconsistency identification, CR: Context-response, S2S: Seq2Seq, Tr.: Transformer, CE: Cross-entropy, HRED: Hierarchical RNN Encoder-Decoder. $\dagger$ Pretrained checkpoints available but only for inference. $\ddag$ Both source-code and checkpoints are not available.}
\label{tab:lit}
\end{table*}

\subsection{Dialog System Pretraining}
There have been quite a few efforts towards utilizing
existing representations or developing new pretrained models for 
dialog systems.
While BERT \cite{devlin2018bert}, ELMo \cite{peters2018deep},
GPT-2 \cite{radford2019language} and other general purpose large-scale pretrained networks are not specific to dialogs, transfer learning from such models could be reasonable. Basic language understanding capability available through these representations
helps to get decent performance on 
many dialog-understanding tasks 
\cite{hosseini2020simple}.


On the other hand, there have been various works on pretraining dialog specific 
representations or large-scale generation models.
We summarize the properties of various previously proposed dialog-representation learning models in Table \ref{tab:lit}.
DialoGPT~\cite{zhang2019dialogpt}, Meena~\cite{adiwardana2020towards} and Blenderbot \cite{roller2020recipes} are large-scale
Transformer-based language models,
which are trained to generate the gold-response (as per the dataset) given a dialog context.
ContextPretrain~\cite{Mehri2019PretrainingMF}, ConveRT \cite{Henderson2020ConveRTEA} and ConvFiT \cite{vulic-etal-2021-ConvFiT} are trained on the response retrieval task
using MultiWoz or Reddit conversations. DEB or Dialog Evaluation using BERT \cite{Sai2020ImprovingDE} is a model based on extended
pretraining of the BERT architecture using Reddit data.  
DialogRPT~\cite{gao2020dialogrpt}, on the other hand, is pretrained to predict human-feedback 
(e.g., upvotes and downvotes) on comments to Reddit threads.
This model is initialized using the weights of DialoGPT model.
\citet{wu2020tod} thoroughly investigate
these existing pretrained representations, both generic and dialog
specific, for understanding their effectiveness on
various goal-oriented dialog-understanding
tasks. 

\subsection{Self-supervised Representation Learning with InfoMax}

Mutual Information maximization (InfoMax) is one of the popular approaches for self-supervised learning, first used by \citet{oord2018infonce} and \citet{belghazi2018mine}.  \citet{oord2018infonce} proposed InfoNCE loss which is an 
estimator for lower bound to mutual information (MI) between 
two continuous-valued random variables. InfoNCE has also been used for other NLP applications like training sentence embeddings (SIMCSE~\cite{gao-etal-2021-simcse}), question answering (QA-InfoMax~\cite{yeh2019qainfomax}), etc. Other estimators for mutual information have also been proposed like MINE (Mutual Information Neural Estimator)~\cite{belghazi2018mine} and SMILE~\cite{song2019limitations}. In general, these estimators are also broadly studied in contrastive Learning (CL) literature for training both self-supervised \cite{mikolov2013efficient,devlin2018bert,liu2019roberta,gao-etal-2021-simcse,Henderson2020ConveRTEA,vulic-etal-2021-ConvFiT} and supervised models \cite{schroff2015facenet,gunel2020supervised}. 
Some prior works in the dialog generation domain have used the concept of mutual information to design loss functions or scoring mechanisms to improve specificity of the generated responses \cite{li2016diversity,yoo2020variational}. These works predominantly used MI either as a regularizer, along with cross entropy loss, or as a scoring function for ranking generated responses in a post-processing step.
In the next section, we derive our pretraining loss function DMI for conversational texts from an information-theoretic perspective. 

\section{Discourse Mutual Information}
\label{sec:dmi-defn-n-loss}

\begin{figure*}[!h]
    \centering
    \includegraphics[width=0.7\textwidth]{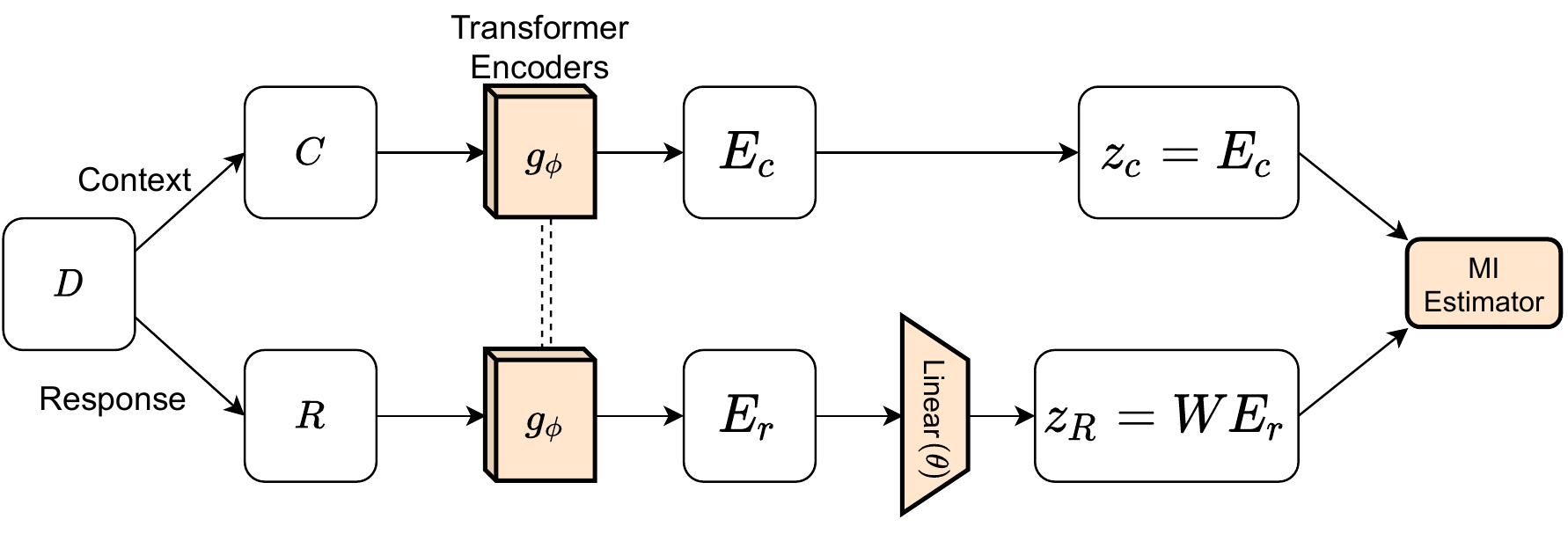}
    \caption{Base Pretraining Architecture for DMI. In our implementations of the model, $f_\phi$ denotes the transformer \cite{vaswani2017attention} based encoders. Context and response encoders share all parameters for efficient learning. $d$ denotes sample dialogs from the training dataset.}
    \label{fig:arch-base}
\end{figure*}

We define Discourse Mutual Information (DMI) as the mutual information\footnote{Mutual Information between two random variables is defined as the reduction
in uncertainty/entropy of one of the random variables by having knowledge 
about the value of the other random variable.
Mathematically, this is written as $I(X;Y) = H(X) - H(X|Y) = H(Y) - H(Y|X)$.} 
between two random variables representing two different segments within the same 
discourse. This is a general concept that can be applied to any form of 
discourse, no matter the domain or type of signal. In this paper, we focus on 
dialog type discourses and representation learning for conversational texts. 
We define two random variables for the contexts ($C$) and 
responses ($R$) that jointly construct a valid conversation. Conversations between humans represent samples from the joint 
distribution $P_{CR}$ of $C$ and $R$. We pose the following learning 
problem, \textit{``learn continuous representations for the textual random variables $C$ and $R$ such that the true mutual information between $C$ and $R$ can be 
closely estimated.''} 

In the remainder of this section we show that, if the lower bound on MI estimated by some representations of context and response is close to the true value, the representation of the context would be as predictive of the response as the natural language form itself.
Existing generative training objectives as used in DialoGPT or Blenderbot are extremely focused on predicting target response only. Per-word cross-entropy loss, used for training these models, fails to take into account the inherent uncertainty in the context-to-response generation function.
Adapting context representations so as to predict the target responses optimally, helps our proposed DMI-based models learn better dialog representations applicable to a versatile set of dialog understanding tasks. 


\paragraph{Objective Function Formulation} 
Let $E_c$ and $E_r$ be the representations\footnote{$C,R,E_c,E_r$ in caps denote the random variables, whereas the lowercased versions $c,r,e_c,e_r$ denote samples.} 
for $C$ and $R$ based on 
some encoder. Using the data processing inequality from Information theory~\cite{cover1999elements}, we have
\begin{align}
    I(C;R) \geq I(E_c; E_r) \label{eqn:data-ineq}
\end{align}
This tells us that 
MI between any encoded version of $C$ and $R$ will always be
less or equal than the true mutual information. 
The equality will hold if $E_c$ and $E_r$ are both 
fully-invertible encoding processes (as opposed to representations which 
are lossy or compressive, and inversion is thus not possible). 
However, neural networks generally embed the data points in a low dimensional manifold by learning robust features that can represent the data points efficiently. Because of this, neural representations are usually not invertible.\footnote{One general exception to this is a neural model/representation overfitted on some training data. In such cases, the model may exactly memorize the input/output pairs.} 
Now, the exact computation of MI is not possible for continuous-valued random variables. In 
recent years, various variational lower bounds have been proposed for 
estimating MI between continuous-valued random variables. Including the MI 
estimator ($\hat I_\theta$), the overall relation becomes
\begin{align}
    I(C;R) \geq I(E_c; E_r) \geq \hat I_\theta (E_c; E_r)
\end{align}

This leads us to the proposed learning objective DMI:
\begin{align}
    \max_{\theta,\phi} \hat I_\theta(E_c^{(\phi)}; E_r^{(\phi)})
\end{align}
where $\hat I_\theta (E_c; E_r)$ is a variational lower bound estimate of $I(E_c; E_r)$ 
(Equation \ref{eqn:data-ineq}) parametrized by $\theta$ and $\phi$ denotes the parameters 
of the encoder used for encoding $C$ (or $R$) to $E_c$ (or $E_r$). 
\paragraph{Loss function} For training our models, we minimize a loss 
function depending on the estimator being used. 
\begin{align*}
  \min_{\theta,\phi} \left[ L_{\theta,\phi}(C,R) = -\hat I_{\theta,estimator}(E_c^{(\phi)}, E_r^{(\phi)}) \right] 
\end{align*}
We experimented with various MI estimators from literature, namely, MINE ~\cite{belghazi2018mine}, InfoNCE ~\cite{oord2018infonce}, JSD~\cite{hjelm2018dim} and SMILE ~\cite{song2019limitations}. These MI estimators generally compute samples of $E_c$ and $E_r$ using $C$ and $R$ drawn from the joint distribution $P_{CR}$. 
Based on our preliminary experiments, we found that InfoNCE estimator produces better representations.
The InfoNCE MI-estimate is computed as,
\begin{align}
    I(C;R) &\geq \log N - L_N \label{eqn:infonce}\\
    L_N &= -\frac{1}{2}\mathbb{E}_{P_{CR}} \left[ \log \frac{e^{f(c,r)}}{\sum_{r'\in R}e^{f(c, r')}} \right]  \nonumber
\end{align}

\noindent where $N$ denotes the batch size, and $f(c,r)$ is a scoring function for the $\langle c,r\rangle$ pair.

\noindent\textbf{InfoNCE-S:}
The original InfoNCE formulation only considers negative samples for one of the random variables, but does not pose any constraint on which of the variables should be considered for negative sampling. As identifying the true response, from a pool of negative samples, would require different reasoning than identifying the true context out of a pool, we consider both these cases and create a symmetric version of the InfoNCE loss function. The final expression of this loss is given in Equation \ref{eqn:infonce-symm} and we refer to it as InfoNCE-S.
This considerably improves the speed of training and convergence, and also gives a boost to downstream task performance. 

\begin{align}
    L_N = -\frac{1}{2}&\mathbb{E}_{P_{CR}} \left[ \log \frac{e^{f(c,r)}}{\sum_{r'\in R}e^{f(c, r')}} \right] \nonumber \\
    &\ \ - \frac{1}{2}\mathbb{E}_{P_{CR}} \left[ \log \frac{e^{f(c,r)}}{\sum_{c'\in C}e^{f(c', r)}} \right]  \label{eqn:infonce-symm}
\end{align}
For other loss functions, more detailed discussion can be found in the Appendix. 

\noindent\textbf{Comparison with ConveRT~\cite{Henderson2020ConveRTEA}}: There are a couple of differences between ConveRT's contrastive loss and our DMI objective. ConveRT models the problem as a response selection task and focuses on modeling cosine similarity between the context and the response. On the other hand, we propose a generic similarity computation function $f(c,r)$ in Eqn. \ref{eqn:infonce} and \ref{eqn:infonce-symm}. Another difference is in encoding the input. ConveRT splits the context into previous turns and current query, and encodes them independently. Our model encodes the entire context jointly and hence is capable of better learning the correlations between previous turns and current query.

\section{DMI vs. Language Modeling Objectives}
In this work, we focus on utilizing DMI for 
pretraining dialog representations incorporating strong discourse-level features. \textit{But why should the DMI objective learn better discourse-level features than models trained on conversational data using MLM or LM objectives?}
We can find the answer by looking at various LM-based objectives through the lens of InfoMax, as shown by \citet{Kong2020AMI}. They connected various pretraining objectives for natural language representations, including the ones used
for training Skipgram, BERT and XLNet, to the InfoMax learning principle.

If we consider an input text $T$ and a masking function $M$ that returns a masked text $\tilde{T}$ and the masked word $w$, the MLM objective is equivalent to $\mathcal{L}_{MLM} = -\hat{I}_\theta(E^{(\phi)}(\tilde{T}), e_w)$ 
where, $E^{(\phi)}$ is the language encoder (e.g., a Transformer encoder) and $e_w$ is the embedding of the token $w$. Similarly, in the case of auto-regressive LMs like GPT-2, the InfoMax objective equivalent to the loss is $\mathcal{L}_{autoLM} = -\hat{I}_\theta(E^{(\phi)}(T_{1:t-1}), e_{T_{t}})$, where $T_{1:t-1}$ is the input sequence till $t-1^{th}$ token and $T_t$ is the $t^{th}$ token.


Compared to these LM objectives, DMI focuses on optimizing $I(E_c, E_r)$, where $c$ and $r$ are two structural components from the discourse with designated roles. This enables DMI to discover more important features at the discourse level.

\section{Experiments}
\label{sec:experiments}

\subsection{Architecture}

The exact encoder architecture and the pretraining pipeline has been shown in Figure \ref{fig:arch-base}. We use a dual encoder architecture for encoding the contexts and responses separately. We observe that sharing parameters between the two encoders leads to a more efficient learning process and faster convergence. We use vanilla transformer-based encoders\footnote{We implemented all models and experiments using the PyTorch and Huggingface libraries.} \cite{vaswani2017attention} for encoding the natural language inputs. The first tokens for both context and response sequences are the special [CLS] tokens whose contextual embeddings from the encoder are used as the context or response representations. The utterances in the context are delimited by another special token [EOU] (for end-of-utterance). We construct the context using as many utterances from the dialog history as possible up to a maximum of 300 subword tokens. We use the WordPiece tokenizer from
BERT for tokenizing the input texts, with a vocabulary size of 30,522. 

The scoring function $f(c, r)$ in Eqs.~\ref{eqn:infonce} and~\ref{eqn:infonce-symm} is implemented using a Bilinear dot product between the context and response representations: $f(c,r) = e_c^T W e_r$ 
where, $W$ is a square weight matrix trained along with other parameters in the model. This function can take any real value, positive or negative, thus allowing the $\hat I_{\theta}(E_C^{(\phi)};E_R^{(\phi)})$ function to take any positive real value. While any complicated function with that range could be chosen, we chose this as a simple formulation satisfying the range constraint and left most of the learning to the transformer and the projection matrix W.

\subsection{Model Variants}

We train three different scales of the DMI model: \textbf{DMI\_Small} with 6 layers, \textbf{DMI\_Medium} with 8 layers, and \textbf{DMI\_Base} with 12 layers. All configurations use 12 attention heads and 768-dimensional embeddings. DMI\_Small is initialized with \textit{``google/bert\_uncased\_L-6\_H-768\_A-12"}\footnote{These pretrained model tags are from the Huggingface model repository.}, DMI\_Medium is initialized with \textit{``google/bert\_uncased\_L-8\_H-768\_A-12"} and DMI\_Base is initialized by  weights from \textit{``RoBERTa-base''} pretrained checkpoint, and further pretrained on the pretraining dataset (see \S\ref{sec:datasets}). All of these models are trained using the InfoNCE-S estimator, unless specified otherwise.


 





\subsection{Hyper-parameter Settings}
We use Adam optimizer with a linear learning rate schedule for training both the models. Learning rate is first linearly increased
to a max value of $5\times 10^{-5}$ during the warm-up phase (first 1000 steps). Following this, in the remaining training period, learning rate is linearly decayed down back to zero. Before training DMI\_Base, we reset the parameters of the 12th self-attention layer, and it is trained again from scratch along with the weight matrix $W$ using our DMI objective. The embedding layer and initial 11 self-attention layers of the RoBERTa-base encoder are finetuned at a slower learning rate ($5\times 10^{-6}$) during our pretraining phase.

As the mutual information value obtained by the InfoNCE loss is upper bounded by $\log(N)$, $N$ being the batch size, we try to keep the value of $N$ as large as possible. Both 8 and 12-layer models are trained on 4-GPU (4x32 GB V100s) systems with overall batch size of 480 and 384, respectively\footnote{Training time: A maximum of 2 weeks of training time was allowed for 8-layer and 12-layer models. Though, the training process saturates long before the maximum allowed time, and we evaluate our models based on checkpoints when the best validation scores are first obtained.}. All the trained models will be publicly shared upon publication.

\begin{table*}
\centering
\scriptsize
\begin{tabular}{@{}llrrrr@{}}
\toprule
\textbf{Task} & \textbf{Description} & \textbf{Train} & \textbf{Valid} & \textbf{Test} & \textbf{Metric} \\ \midrule
Banking77 & Intent 77-class Classification & 8,002 & 2,001 & 3,080 & Accuracy \\
SWDA & Dialog Act 41-class Classification & 213,543 & 56,729 & 4,514 & Accuracy \\
MuTual & Reasoning as Response Selection & 25,516 & 2,836 & 3,544 & R@1, R@2, MRR \\
MuTual Plus & MuTual + Safe response candidate & 25,516 & 2,836 & 3,544 & R@1, R@2, MRR \\
DD++ & Dialog Evaluation & 92,590 & 10,280 & 11,420 & Accuracy \\
DD++/Adv & Train: Adv. neg., Test: Adv neg. samples & 92,590 & 10,280 & 11,420 & Accuracy \\
DD++/Cross & Train: Random neg., Test: Adv neg. samples & 92,590 & 10,280 & 11,420 & Accuracy \\
DD++/Full & Train: All samples, Test: Adv. neg. samples & 138,885 & 10,280 & 11,420 & Accuracy \\
Empathetic Intent & Emotion and Intent 44-class Classification & 25,023 & 3,544 & 3,225 & Accuracy \\ 
\bottomrule
\end{tabular}%
\caption{Downstream task details. Adv.: Adversarial, Neg.: Negative}
\label{tab:downstream}
\end{table*}

\subsection{Pretraining Dataset}
\label{sec:datasets}
We pretrained all our models using the Reddit corpus (\textbf{Reddit-727M conversational-data}) released by \citealp{henderson2019repository}. We ran the scripts released by the authors to recreate the dataset of 727M English conversations. Out of these 727M conversations, we utilize around 7.5\% to 10\% of the dataset to train our models, after which the validation loss generally saturates. In the rest of this paper, we will refer to this dataset as \textbf{rMax}, in short.

\paragraph{Dialog Unrolling for Pretraining}
For training our models, we need samples of context-response (CR) pairs. 
Each dialog is 
unrolled to create context-response pairs with each utterance in the dialog 
as a response, except the first one. Hence, for each dialog $D=\{U_1, U_2, \dots, U_T\}$, we generate the following 
set of samples $S=\{(\mathcal{C}_t: U_1, \dots, U_{t-1}; \mathcal{R}_t: U_{t}):t\in[2, T]\}$. If we process the full rMax dataset, this leads to, approximately, 2.7B CR pairs.

\subsection{MI Estimation}
\label{sec:exp-mi-estimation}
During pretraining, we compare the checkpoints from different epochs and across hyperparameter settings in terms of the bits of mutual information extracted by the trained representation on an unseen set of dialogs. This is calculated as $MI_{valid}=\log(N) - L_N$ (see \S\ref{sec:dmi-defn-n-loss} for more details).
As per the Information Bottleneck theory \cite{tishby2000information}, the mutual information learned between the two observed random variables can be factorized into two components, namely, predictive and redundant information. Predictive information generally identifies whether the features learned by the representation are useful for a downstream task. 
The redundant information is caused by features that do not help in any downstream tasks. Such features can exist due to noise or spurious correlation in the dataset, or even overfitting. Hence, we train our final models on a fraction of the rMax dataset but only for one epoch (i.e., we never repeat the samples) which removes any possibility of overfitting.

Predictive features identified based on a fixed set of downstream tasks \cite{tishby2000information,alemi2016deep} may not be a sufficient to assess other features learned in the training process. Since, ideally, we want to maximize the amount of predictive information in the representation, we compare the bits of MI on the training set against the bits of MI on an unseen validation set, as captured by the learned representation. To make sure that we do not assume anything about the domain or the conversation topics, we use the validation set of dialogs from the open-domain Daily Dialog dataset~\cite{li2017dailydialog}.

\subsection{Downstream Tasks}

Instead of focusing on a single downstream task like many previous works on dialog representation learning, we consider a more versatile range of tasks to evaluate the learned representations from DMI or the baseline models.
To find out whether a certain representation is effective for some downstream task, we evaluate in two setups: probe and finetune. In both cases, the pretrained model is used along with an MLP classifier of fixed complexity \cite{pimentel2020information}. In probing setup, we only train the parameters of the MLP classifier. In finetuning setup, we also train the pretrained model parameters along with the MLP classifier parameters. We use the context and response representations from our models as the input to the MLP classifier. 

For downstream tasks, we have two reasoning tasks 
based on the MuTual dataset \cite{cui2020mutual}, three classification 
tasks based on conversational intent detection~\cite{casanueva2020efficient},  emotion detection \cite{welivita2020taxonomy} and act classification \cite{stolcke1998dialog}, and four 
dialog evaluation tasks based on the 
DailyDialog++ dataset (DD++, \citealp{Sai2020ImprovingDE})\footnote{Note that DailyDialog++ is different from DailyDialog.}. Table \ref{tab:downstream} shows dataset details and metrics for these nine tasks. Both MuTual and DailyDialog++ datasets have an adversarial configuration for the respective tasks, which allows us to assess each of the evaluated models in adversarial settings also.

\section{Results and Discussions}


\setlength{\tabcolsep}{4.5pt}
\begin{table*}[!thb]
\centering
\scriptsize
\begin{tabular}{@{}llrrrrrrrrrrrrr@{}}
\toprule
&& B77 & SWDA & E-Intent& \multicolumn{3}{c}{MuTual} & \multicolumn{3}{c}{MuTual Plus} & DD++ & DD++/adv & DD++/cross & DD++/full \\ 
\midrule
&Model & Acc. & Acc. & Acc. & R@1 & R@2 & MRR & R@1 & R@2 & MRR  & Acc. & Acc. & Acc. & Acc. \\\midrule
\multirow{12}{*}{\rotatebox{90}{Probing}}&RoBERTa\_Base & 72.84 & 67.18 & 50.45 & \textbf{49.70} & \textbf{75.20} & \textbf{70.00} & 43.60 & 66.60 & 65.10 & 55.75 & 84.20 & 65.11 & 68.76 \\
&BERT\_Base & 72.74 & 67.99 & 46.84 & 45.40 & 72.80 & 67.30 & 42.60 & 67.70 & 64.90 & 60.39 & 86.56 & 65.25 & 72.50 \\
&T5\_Base & 60.82 & 68.79 & 44.50 & 43.20 & 69.40 & 65.60 & 38.30 & 65.70 & 62.20 & 57.46 & 84.14 & 61.23 & 63.35 \\
&GPT-2\_Small & 76.64 & 69.17 & 49.94 & 44.92 & 70.54 & 66.60 & 40.75 & 66.70 & 63.46 & 67.37 & 82.06 & 67.53 & 73.93 \\
&DialoGPT\_Small & 53.00 & 65.10 & 43.42 & 29.80 & 53.50 & 55.15 & 25.51 & 57.56 & 54.05 & 63.63 & 78.02 & \textbf{70.61} & 70.77 \\
&Blender\_Small & 70.39 & 70.11 & 48.52 & 41.42 & 68.06 & 64.29 & 42.89 & 68.85 & 65.18 & 60.07 & 65.14 & 57.76 & 68.20 \\
&ConveRT & \textbf{89.88} & \textbf{71.36} & \textbf{55.47} & 45.30 & 72.00 & 67.00 & 40.90 & \textbf{69.00} & 64.30 & \textbf{79.14} & \textbf{88.67} & 69.59 & \textbf{80.86} \\
&DialogRPT & 81.54 & 67.92 & 50.74 & 39.50 & 66.80 & 63.00 & 34.20 & 61.50 & 59.20 & 74.11 & 81.29 & 68.49 & 67.20 \\
&DEB & 79.18 & 68.50 & 45.31 & 45.10 & 74.00 & 67.50 & \textbf{45.00} & 67.70 & \textbf{66.00} & 70.66 & 86.07 & 67.25 & 67.77 \\
\cmidrule{2-15}
& DMI\_Small & 89.81 & 72.33 & 55.72 & 51.24 & 74.94 & 70.76 & 46.39 & 70.09 & 67.14 & 85.01 & 91.51 & 75.75 & 86.34 \\
& DMI\_Medium & 90.42 & 72.49 & 57.33 & 52.48 & 75.62 & 71.47 & 46.61 & 72.46 & 67.79 & 85.80 & 91.38 & 76.73 & 86.94 \\
& DMI\_Base & 91.43 & 72.73 & 60.00 & 52.48 & 76.41 & 71.65 & 48.98 & 71.33 & 68.73 & 86.91 & 91.98 & 79.15 & 88.32 \\
\cmidrule{2-15}
&$\Delta$ & \textcolor{OliveGreen}{1.55}&\textcolor{OliveGreen}{1.37}&\textcolor{OliveGreen}{4.53}&\textcolor{OliveGreen}{2.78}&\textcolor{OliveGreen}{1.21}&\textcolor{OliveGreen}{1.65}&\textcolor{OliveGreen}{3.98}&\textcolor{OliveGreen}{2.33}&\textcolor{OliveGreen}{2.73}&\textcolor{OliveGreen}{7.77}&\textcolor{OliveGreen}{3.31}&\textcolor{OliveGreen}{8.54}&\textcolor{OliveGreen}{7.46} \\
\midrule
\midrule
\multirow{11}{*}{\rotatebox{90}{Finetuning}}&RoBERTa\_Base & \textbf{92.75} & 73.61 & \textbf{62.81} & 48.42 & \textbf{77.20} & 69.70 & \textbf{49.55} & \textbf{73.70} & \textbf{69.50} & 90.00 & 95.70 & \textbf{73.76} & 91.09 \\
&BERT\_Base & 92.27 & 72.29 & 60.12 & 47.86 & 73.93 & 68.80 & 49.10 & 72.35 & 69.00 & 87.05 & 94.33 & 67.70 & 88.82 \\
&T5 & 89.11 & \textbf{73.77} & 60.66 & 49.77 & 73.93 & 69.80 & 43.00 & 66.93 & 64.90 & 82.03 & 90.89 & 65.85 & 85.63 \\
&GPT-2\_Small & 92.49 & 72.62 & 58.44 & 48.42 & 72.69 & 68.90 & 45.71 & 70.99 & 67.10 & 85.69 & 93.60 & 68.43 & 87.83 \\
&DialoGPT\_Small & 92.59 & 73.48 & 59.33 & 49.32 & 75.17 & 69.80 & 47.86 & 73.02 & 68.44 & 83.68 & 91.99 & 64.06 & 85.54 \\
&Blender\_Small & 91.59 & 71.10 & 58.31 & \textbf{52.93} & 75.85 & \textbf{71.80} & 47.97 & 70.99 & 68.30 & 86.83 & 92.29 & 66.39 & 87.82 \\
&DialogRPT & 92.70 & 72.02 & 62.13 & 52.14 & 76.19 & 71.40 & 46.95 & 70.54 & 67.66 & \textbf{90.26} & \textbf{95.81} & 73.34 & \textbf{91.25} \\
&DEB & 92.53 & 72.14 & 59.69 & 48.19 & 74.49 & 69.00 & 46.95 & 70.65 & 67.80 & 85.74 & 94.05 & 64.42 & 89.02 \\
\cmidrule{2-15}
& DMI\_Small & 92.44 & 71.29 & 61.05 & 55.42 & 75.28 & 72.92 & 47.63 & 72.01 & 68.19 & 87.57 & 94.99 & 77.33 & 88.96 \\
& DMI\_Medium & 92.76 & 71.53 & 62.88 & 55.76 & 77.88 & 73.56 & 50.68 & 73.25 & 70.04 & 89.12 & 95.63 & 78.26 & 90.80 \\
& DMI\_Base & 93.93 & 74.50 & 64.62 & 56.43 & 79.91 & 74.27 & 52.14 & 75.06 & 71.09 & 91.03 & 96.39 & 81.69 & 92.61 \\
\cmidrule{2-15}
&$\Delta$ & \textcolor{OliveGreen}{1.18} & \textcolor{OliveGreen}{0.73} & \textcolor{OliveGreen}{1.81} & \textcolor{OliveGreen}{3.50} & \textcolor{OliveGreen}{2.71} & \textcolor{OliveGreen}{2.47} & \textcolor{OliveGreen}{2.59} & \textcolor{OliveGreen}{1.36} & \textcolor{OliveGreen}{1.59} & \textcolor{OliveGreen}{0.77} & \textcolor{OliveGreen}{0.59} & \textcolor{OliveGreen}{7.93} & \textcolor{OliveGreen}{1.35} \\ 
\bottomrule
\end{tabular}%
\caption{Results from probing (top) and finetuning (bottom) setups on 9 downstream tasks for assessing dialog
understanding. (DD++: DailyDialog++, B77: Banking77 task, R@k: Recall at k, MRR: Mean reciprocal rank). Our model consistently performs better than SOTA on all the tasks in both probing as well as finetuning setups.}
\label{tab:results}
\end{table*}

\begin{table*}[!thb]
\centering
\scriptsize
\begin{tabular}{@{}lrrrrrrrrrrrrr@{}}
\toprule
& B77 & SWDA & E-Intent& \multicolumn{3}{c}{MuTual} & \multicolumn{3}{c}{MuTual Plus} & DD++ & DD++/adv & DD++/cross & DD++/full \\ 
\midrule
Model & Acc. & Acc. & Acc. & R@1 & R@2 & MRR & R@1 & R@2 & MRR  & Acc. & Acc. & Acc. & Acc. \\
\midrule
DMI\_Base & \textbf{93.93} & \textbf{74.50} & 64.62 & 56.43 & \textbf{79.91} & 74.27 & \textbf{52.14} & \textbf{75.06} & \textbf{71.09} & \textbf{91.03} & 96.39 & 76.01 & 92.61 \\
DMI\_Base - Sym & 93.28 & 72.69 & \textbf{65.18} & \textbf{57.34} & 77.88 & \textbf{74.32} & 48.08 & 72.69 & 68.60 & 90.94 & \textbf{96.65} & \textbf{76.45 }& \textbf{93.13} \\
DMI\_Base - RoB & 92.34 & 74.10 & 60.96 & 53.84 & 77.31 & 72.47 & 50.34 & 72.80 & 69.81 & 87.23 & 92.95 & 73.53 & 87.85 \\
DMI\_Base - Sym - RoB & 91.59 & 73.55 & 60.71 & 54.06 & 75.40 & 72.24 & 47.97 & 71.45 & 68.24 & 86.79 & 92.96 & 70.29 & 87.13 \\
\bottomrule
\end{tabular}%
\caption{Ablation study results for the finetune setup for our base model on 9 downstream tasks. ``-RoB'' $\rightarrow$ No RoBERTa initialization. ``-Sym'' $\rightarrow$ Training with non-symmetric version of InfoNCE. (DD++: DailyDialog++, B77: Banking77 task, R@k: Recall at k, MRR: Mean reciprocal rank).  
}
\label{tab:ablations}
\end{table*}

\subsection{Pretraining DMI based Representations}

During pretraining, we used ``Validation MI'' to evaluate model checkpoints. As the goal of our models is to learn a representation that captures maximum MI between the context and the response texts, this metric tracks 
how well the learned representation captures the mutual information between contexts and responses of unseen dialogs.


We use the validation split from Daily Dialog dataset as our validation set for evaluation the model during pretraining. It is not specific to a domain and, hence, covers a versatile range of topics.  
This set comprises 1,000 full conversations between two persons which on unrolling leads to 7,069 context-response (CR) pairs. We illustrate the variation in validation-MI metric against training steps in Fig. \ref{fig:pretrain-mi} in the Appendix.

\subsection{Comparison of Representations on Downstream Task Performance}

In this set of experiments, 
we probe/finetune the DMI models with 
various downstream tasks that require knowledge of many different types
of dialog-understanding features. The results of our probing and finetuning experiments
are shown in Table~\ref{tab:results}.


We have used two types of models as our baselines: generic pretrained models and dialog-specific pretrained models. RoBERTa, BERT, T5~\cite{raffel2019exploring}, GPT-2 are all trained on large corpora of generic web-crawled English text. But, since these models were not specifically pretrained on any dialog corpus, they may suffer from poor performance on certain dialog understanding tasks. Hence, we consider DialoGPT\footnote{DialoGPT and DEB are based on GPT-2 and BERT models and were further pretrained on conversations from Reddit. They use the original loss functions of GPT-2/BERT.}, DialogRPT, DEB and ConveRT models, which were trained on conversational data. For DialogRPT, we used ``human-vs-rand'' checkpoint released by authors. All models are 12-layer except Blender\_Small (8 layers), ConveRT (6 layers), DialogRPT (24 layers) and DMI\_Medium (8 layers). We used the publicly available model checkpoints for all baselines, wherever possible. 
The ConveRT model's checkpoint has been removed from Github\footnote{\url{https://github.com/PolyAI-LDN/polyai-models}} by its authors. Hence, it was only possible for us to MLP-probe the representations, without finetuning of the model, based on a cached version released by another user under a valid license\footnote{\url{https://github.com/davidalami/ConveRT}}. Pretrained checkpoints for Meena, ContextPretrain and ConvFiT are not available, and hence we do not compare with them.

\subsubsection{Results in Probing Setup}

We observe that, on average, DEB and ConveRT have good performance among the baselines. However, the RoBERTa model outperforms all other baselines on the MuTual task by a significant margin. In the MuTual Plus task, the DEB model outperforms other models in the R@1 and MRR metrics.
ConveRT performs the best among all baselines on the other tasks. ConveRT's loss function is also contrastive in nature and is similar to ours. This explains the model's generally high strength across the tasks among all the baselines. 

Our DMI\_Base beats ConveRT on all the tasks, and DMI\_Medium beats the baseline on 7 out of 9 tasks.
We believe DD++ tasks to be the most demanding ones with respect to context-level understanding. Here, all non-dialog baselines have a weaker performance, with DEB and ConveRT being the best of the bunch. These are also the tasks where our models excel the most, with both DMI\_Medium and DMI\_Base beating all baselines with strong margins. DD++/cross is the most difficult among all four DD++ tasks. Here, the model is trained on random negative samples and tested on a dataset with human-curated adversarial negatives. 
Our DMI\_Base beats the best baseline on DD++/cross by 8.54 points. This shows the superior quality of context representations from our models.


\subsubsection{Results in Finetuning Setup}

In the finetuning setup, on average, RoBERTa and DialogRPT have good performance among the baselines. DialogRPT performs well for DD++ tasks while Blender works well for the MuTual task. For all other tasks, RoBERTa is the best baseline, even outperforming models especially trained for dialog tasks (like DialoGPT). 

Similar to the probe setup, DMI\_Base beats baseline methods by significant margins. In general, finetune results are better than probe results across all models, as expected. 

Our large-scale RoBERTa-initialized DMI\_Base model outperforms the best baseline for all tasks, by a considerable margin.
Additionally, our DMI-based models are able to perform well uniformly across all tasks, unlike even baselines like DialoGPT, DialogRPT and Blenderbot models which are explicitly trained on dialog data.  
This makes DMI the best overall model for dialog related tasks. Across multiple tasks, we show qualitative examples where our proposed DMI-based models provide accurate results, in the Appendix.

\subsection{Ablations} 

We evaluate the importance of using RoBERTa based pretraining as well as the symmetric version of the InfoNCE loss in Table~\ref{tab:ablations}. We observe that RoBERTa based pretraining helps significantly across all tasks. The symmetric InfoNCE improves performance for SWDA and MuTual Plus tasks.


\section{Conclusions and Future work}

In this paper, we proposed the concept of Discourse Mutual Information (DMI) which is better suited for learning dialog-specific features in a self-supervised manner.
Using the InfoMax principle we formulated a pretraining method for dialog-specific representation learning. 
Across 9 downstream dialog understanding tasks, our 12-layer model outperforms state-of-the-art methods. Further, we showed that on most of these tasks, even our 8-layer model outperforms standard 12-layer pretrained models. These experiments show the potential of the proposed DMI objective towards building dialog understanding models. We will make the code and pretrained model checkpoints available on request, instructions can be found here \url{https://bsantraigi.github.io/DMI}. Although we experimented only with dialog modeling in this paper, we believe that the proposed DMI objective is generic enough to be applied to any type of discourse in any domain.
In the future, we would like to explore how to harness DMI representations for generative conversation modeling.
\section{Acknowledgments}
This work was partially supported by Microsoft Academic Partnership Grant (MAPG) 2021. The first author was also supported by Prime Minister's Research Fellowship (PMRF), India.

\section{Ethical considerations}

Like many other pretrained language representation models, the proposed model may also have learned patterns associated with exposure bias. Interpretability associated with the output is rather limited, hence users should use the outputs carefully. The proposed model ranks possible response candidates, and does not filter out any ``problematic'' candidates. Thus, for applications, where candidate responses could be problematic, (e.g., offensive, hateful, abusive, etc.), users should carefully filter them out before providing them as input to our model. 

All the datasets used in this work are publicly available. We did not collect any new dataset as part of this work.

Banking77~\citealp{casanueva2020efficient} has been obtained from \url{https://github.com/PolyAI-LDN/task-specific-datasets}. It is available under a Creative Commons Attribution 4.0 International license with details here\footnote{\url{https://github.com/PolyAI-LDN/task-specific-datasets/blob/master/LICENSE}}.

SWDA~\citealp{stolcke1998dialog}: The dataset has been obtained from \url{http://compprag.christopherpotts.net/swda.html}. This work is licensed under a Creative Commons Attribution-NonCommercial-ShareAlike 3.0 Unported License.

E-Intent~\citealp{welivita2020taxonomy}: The dataset was downloaded from \url{https://github.com/anuradha1992/EmpatheticIntents}. The original dataset is available at ~\url{https://github.com/facebookresearch/EmpatheticDialogues} which is under the Creative Commons Attribution 4.0 International license.

MuTual and MuTual-plus~\citealp{cui2020mutual}: The datasets have been downloaded from ~\url{https://github.com/Nealcly/MuTual}. Licensing is unclear; the authors do not mention any license information or terms of use.

DailyDialog++~\citealp{Sai2020ImprovingDE}: The dataset was downloaded from \url{https://github.com/iitmnlp/DailyDialog-plusplus}. The data is available under the MIT License.

rMax or Reddit-727M conversational-data~\citealp{henderson2019repository}: the dataset has been obtained from \url{https://github.com/PolyAI-LDN/conversational-datasets/tree/master/reddit}. The dataset is available under the Apache License Version 2.0.

\bibliography{main-smi}
\bibliographystyle{acl_natbib}

\appendix

\section{Mutual Information Estimators}



In this paper, we experiment with various different MI estimators, and found InfoNCE-S to be the best (both in terms of accuracy as well as training speed). 
The mathematical formulation of these estimators is provided below.
\begin{enumerate}
    \item InfoNCE was proposed by \citet{oord2018infonce}. It connects to the mutual information value $I(X;Y)$ as,
    \begin{align*}
        I(X;Y) &\geq \log(N) - L_N\\
        L_N &= -\mathbb{E}_{P_{XY}} \left[ \log \frac{e^{f(x,y)}}{\sum_{y'\in Y}e^{f(x, y')}} \right] \label{eqn:apndx-infonce}
    \end{align*}
    \item MINE \cite{belghazi2018mine}
    \begin{align*}
    I(&X;Y) \geq sup_{\theta\in \Theta} \big[ I_{\theta}^{(MINE)}(X;Y) =\\
    &\mathbb{E}_{P_{XY}} [T(x, y)] - \log \mathbb{E}_{P_X\times P_Y} [e^{T(x,y)}]  \big]
    \end{align*}
    \item JSD \cite{hjelm2018dim}
    \begin{align*}
      I^{(JSD)}_\theta (X;Y) &= \mathbb{E}_{P_{XY}} [-sp(-T(x, y))] \\
      &- \mathbb{E}_{P_X\times P_Y} [sp(T(x,y))]
    \end{align*}

    \item SMILE \cite{song2019limitations}
    \begin{align*}
    I_{\theta}^{(smile)}&(X;Y) = \mathbb{E}_{P_{XY}} [T(x, y)] \\
    &- \log \mathbb{E}_{P_X\times P_Y} [clip(e^{T(x,y)}, e^{-\tau}, e^\tau)] 
    \end{align*}

\end{enumerate}

Use of the InfoMax objective for self-supervised learning has been more prevalent in the computer vision domain than in NLP. Although as \citet{Kong2020AMI} have previously shown, many existing loss functions used for training NLP models can be derived directly from the InfoMax framework. \citet{Kong2020AMI} had only focused on various language model objectives that focus on words given the surrounding context. The authors showed that this objective translates to maximizing mutual information between the context and the missing word within the context. 

In dialog domain also, InfoMax-equivalent loss functions have been used.
First, \citet{Henderson2020ConveRTEA} used contrastive formulation of the response selection task as a pretraining objective for dialog representation. Other prior works on response selection models often used a binary-cross entropy loss for training. Both these loss functions are actually equivalent to various lower bound estimators for mutual information.
In the QAInfoMax model \cite{yeh2019qainfomax}, the authors used the Deep-InfoMax loss function \cite{hjelm2018dim} as a regularizer and showed that representations learned with or in-presence of an InfoMax regularizer are more resilient to adversarial attacks while maintaining the same level of task performance. We also observe the same effect in our DD++/cross experiments. This is because of the self-supervised yet task-specific nature of the loss function. 



\begin{figure}[h!]
  \centering
  \includegraphics[width=0.85\columnwidth]{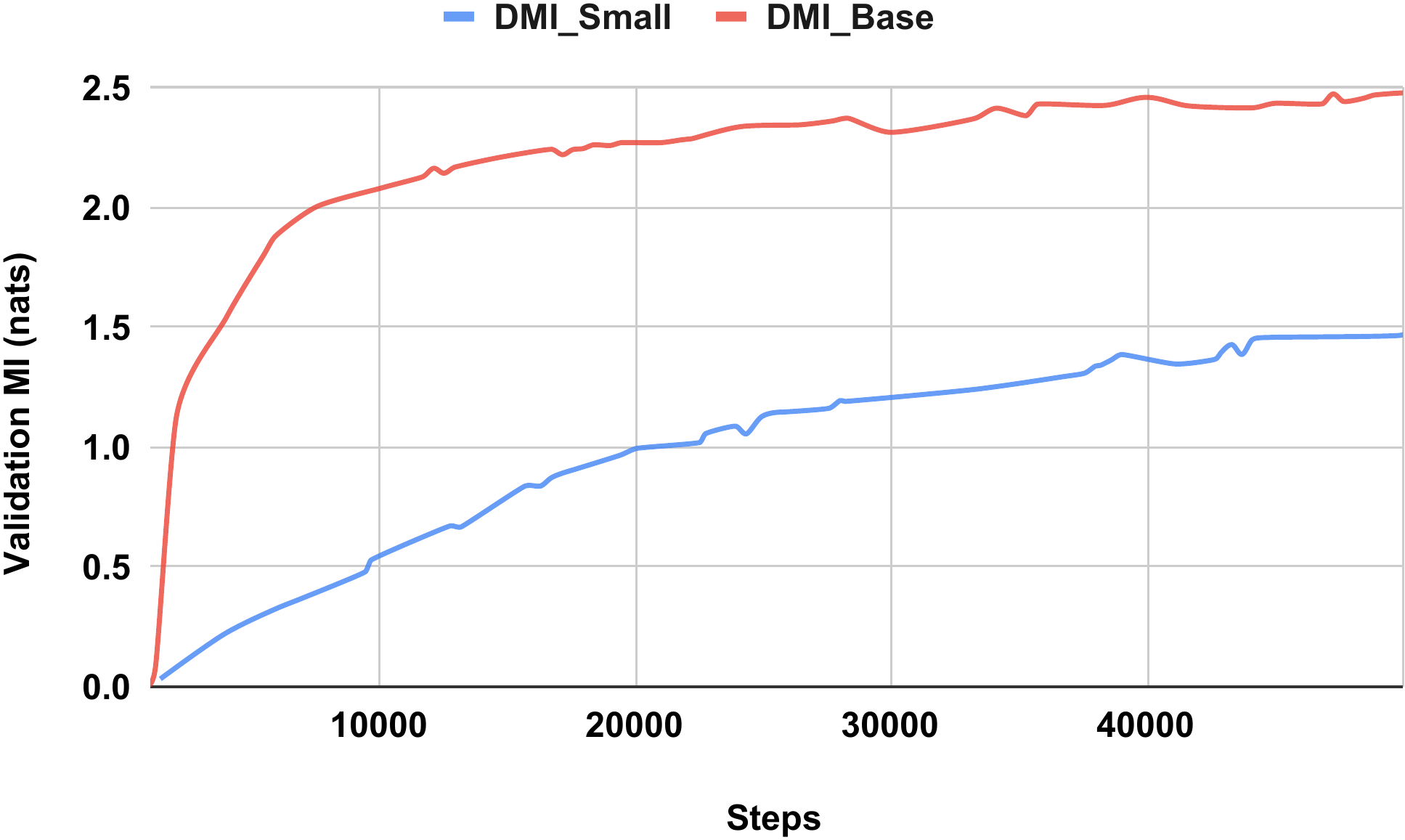}
  \caption{Validation-MI profile during pretraining}
  \label{fig:pretrain-mi}
\end{figure}

\section{Response Retrieval Experiment}
We wanted to investigate if the proposed model can rank good responses higher compared to more generic/bland ones. 
Hence to test against an extreme setting we simulate a response selection task for a very large pool using the test set of Daily Dialog~\cite{li2017dailydialog} dataset. We took all the $\sim$7000 responses from test set of the daily dialog dataset as the response pool. Next, for a few randomly selected context examples, we illustrate the top two ranked as well as ground truth responses for two full conversations in Tables~\ref{tab:retrieval1} and~\ref{tab:retrieval2}. Of course, the ground truth response was removed from the pool for each context. The ranking of responses were done using the $f(c,r)$ function from the trained DMI\_Base model. From the examples of response selection, we can observe that the model is able to both avoid blend responses and select responses that are relevant to the current context even from such a large pool. This shows the usefulness of dialog specific pretrained representation trained using the DMI objective.

\begin{table*}
    \centering
    \scriptsize
    \begin{tabular}{|p{2in}|p{1.7in}|p{1.7in}|}
\hline
Dialog&Rank-1 Response&Rank-2 Response\\
\hline
\hline
\textbf{S1:} Hey man , you wanna buy some weed ? &Yes , smoking or non-smoking ?&Sure , what ’ s up ?\\
\hline
\textbf{S2:} Some what ? &My cigarettes .&Come on man ! I even got dope and acid ! Try some !\\
\hline
\textbf{S1:} Weed ! You know ? Pot , Ganja , Mary Jane some chronic ! &I'm sorry , Mrs . Lin . We don't allow smoking in the office .&Ha ! That's right . You used to smoke . I forgot !\\
\hline
\textbf{S2:} Oh , umm , no thanks . &Come on man ! I even got dope and acid ! Try some !&Come on , you can at least try a little , besides your cigarette .\\
\hline
\textbf{S1:} I also have blow if you prefer to do a few lines . &No , thanks . Oh , yes , could I have some juice for the mini bar ?&No , thanks . I ' Ve had enough . I'll have my bill , please .\\
\hline
\textbf{S2:} No , I am ok , really . &Great ! Would you prefer smoking or nonsmoking ?&Come on , you can at least try a little , besides your cigarette .\\
\hline
\textbf{S1:} Come on man ! I even got dope and acid ! Try some ! &I will , but let's go to the disco after dinner .&No , thanks . I've had more than enough.In fact , I must be running along .\\
\hline
\textbf{S2:} Do you really have all of these drugs ? Where do you get them from ? &I found them at JC Pennies .&They are my personal effects .\\
\hline
\textbf{S1:} I got my connections ! Just tell me what you want and I ’ ll even give you one ounce for free . &Hmm ... I'll tell you what . I'll lend you four thousand dollars , but you have to pay me back next week .&No , thanks . Oh , yes , could I have some juice for the mini bar ?\\
\hline
\textbf{S2:} Sounds good ! Let ’ s see , I want . &Here you go.On any other day , it would cost me a fortune , but it ’ s on special offer today .&All right , Maria . I ’ ll give you until tomorrow at 4:00 to produce a satisfactory piece of work , but otherwise , you ’ ll have to re-do it .\\
\hline
\textbf{S1:} Yeah ?&But of course ! Well , it ’ s been great talking to you , but I have to get going .&Good . I was thinking that I ’ d like to invite you to watch a movie . I can meet you at the cinema gate .\\
\hline
    \end{tabular}
    \caption{Retrieval Example 1: Top two ranked responses from a large pool, as well as ground truth response for a conversation. Note that every line corresponds to one utterance in the conversation either from speaker S1 or speaker S2. \textbf{How to read the table:} For any context (all entries in first column upto any row), the ground truth response can be automatically obtained from the immediate next entry in first column. The response selected by the model in any cell $(t, 1 \text{ or } 2)$ is for the true context from row $1$ to row $t$ in the first column.}
    \label{tab:retrieval1}
\end{table*}

\begin{table*}
    \centering
   \scriptsize
    \begin{tabular}{|p{2in}|p{1.7in}|p{1.7in}|}
\hline
Dialog&Rank-1 Response&Rank-2 Response\\
\hline
\hline
\textbf{S1:} Could I have my bill , please ? & Sure . Here is your receipt . & Sure . Your cash back is \$ 13 . And we'll bring out your fries in two minutes .\\ \hline
\textbf{S2:} Certainly , sir . &  Thanks . Now can I make the full deposit ? & Thank you ! Would you like an aisle seat or a window seat ?\\ \hline
\textbf{S1:} I'm afraid there's been a mistake . &  I am really sorry too , maybe I can give you a call sometime . & Oh , I'm sorry.However , if you could help me out , I'll double the pay for the hours worked .\\ \hline
\textbf{S2:} I'm sorry , sir . What seems to be the trouble ? &  Not much . I had to pay an unexpected bill , so I needed the money back . & Oh , nothing special . I'm just a bit tired .\\ \hline
\textbf{S1:} I believe you have charged me twice for the same thing . Look , the figure of 6.5 dollar appears here , then again here . &  One moment , please , sir . ... Here's your bill . Would you like to check and see if the amount is correct ? & Sir , I deleted the \$ 10 , but I had to add a \$ 2 service charge to your bill.\\ \hline
\hline
    \end{tabular}
    \caption{Retrieval Example 2: Top two ranked responses from a large pool, as well as ground truth response for a conversation. Note that every line corresponds to one utterance in the conversation either from speaker S1 or speaker S2. \textbf{How to read the table:} For any context (all entries in first column upto any row), the ground truth response can be automatically obtained from the immediate next entry in first column. The response selected by the model in any cell $(t, 1 \text{ or } 2)$ is for the true context from row $1$ to row $t$ in the first column.}
    \label{tab:retrieval2}
\end{table*}


\section{Prediction Samples and Error Analysis}

\begin{table*}
\scriptsize
\centering
\begin{tabular}{@{}rL{6.5cm}L{3.5cm}rrr@{}}
\toprule
\textbf{ID} &\textbf{Context} & \textbf{Candidate Response} & \textbf{Gold} & \textbf{DialoGPT} & \textbf{DMI\_Base} \\ \midrule
\multicolumn{5}{c}{\textit{DMI\_Base predicts correctly}} \\ \midrule
1 & All right. I'll take it. \_\_eou\_\_ Do you like to use chopsticks \_\_eou\_\_ Yes, I like using chopsticks. & When you get closer, you see that each horizontal section is made up of two pieces that converge in a right angle. & 0 & 1 & 0 \\ \midrule
2 & And you'll have to sell your motorcycle. And your cameras. Right? \_\_eou\_\_ Maybe I'll cook once or twice a week. How is that? & I go to the temple twice a week so I prefer vegetarian food. & 0 & 1 & 0 \\ \midrule
3 & But I heard the box office rose up to 15 million in the first week. \_\_eou\_\_ Box office can't explain everything. I do not think it is cheerful or well-made. The plot is old and the female character is not pretty. \_\_eou\_\_ My sister has given me two tickets for tonight. It is called' The life of Rose', a French movie. & I got 1 million views on my youtube channel in one week. & 0 & 1 & 0 \\ \midrule
4 & Glad you like it. By the way, is this your first time to China, Mr. White? \_\_eou\_\_ Yes, as a representative of IBM. I hope to conclude some business with you. \_\_eou\_\_ We also hope to expand our business with you. & May I know what and all process you have? & 1 & 0 & 1 \\ \midrule
5 & Good. I have to go right now. I really hope this meeting doesn ’ t last too long. \_\_eou\_\_ They usually go on for ages. \_\_eou\_\_ I ’ ll stop by if I have time later. Make sure everyone knows that we must stick to the deadlines. & I don't cut my hair because I really like to keep it long. & 0 & 1 & 0 \\ \midrule
6 & Of course. The main thing is that all our work must be completed on schedule. We even allow our employee to go home early if they finish their work early. \_\_eou\_\_ How often do you have meetings? \_\_eou\_\_ You should attend a department meeting every Monday morning. There are other meetings for people working together on certain projects. Department heads also attend an interdepartmental meeting each week. & In the newsletter, I gave employees column references this week. & 0 & 1 & 0 \\ \midrule
7 & Sounds interesting! That must be very convenient. \_\_eou\_\_ Yes, you're right. I can blog wherever and whenever I'm on the move. It's especially good when I'm on a business trip and my laptop happens to be away from me. \_\_eou\_\_ How can you do that? & I sank parents money into my business it is not convenient. & 0 & 1 & 0 \\ \midrule
8 & There is a wait right now to use the computers. \_\_eou\_\_ That ’ s fine. \_\_eou\_\_ Would you please write your name on this list? & Sure, please give me a pen. & 1 & 0 & 1 \\ \midrule
\multicolumn{5}{c}{\textit{DMI\_Base predicts wrongly}} \\ \midrule
9 & How much cash would you like? \_\_eou\_\_ I want \$150. \_\_eou\_\_ Here ’ s your \$150. & Well! I never forget your help. & 1 & 0 & 0 \\ \midrule
10 & I see, sir. This one is very good. \_\_eou\_\_ Is it? \_\_eou\_\_ You may rest assured. It sells well. \_\_eou\_\_ May I have a look at the introduction? & It has been recommended by top nutritionists. & 1 & 0 & 0 \\ \midrule
11 & Sir, tell us about your experience with Super Bulk-up. \_\_eou\_\_ Well, it's completely changed my life. \_\_eou\_\_ Tell us how. & The change is right in front of you, isn’t it? & 1 & 0 & 0 \\ \bottomrule
\end{tabular}
\caption{Sample Predictions from the DD++/Cross task. In DD++/Cross, the models are trained using randomly sampled negatives and tested on curated adversarial negative samples. In each sample, the input context comprises the utterances, previous to the response, spoken by the two participants. Such utterances within a context are delimited by a special token ``\_\_eou\_\_''.}
\label{tab:dd-cross-samples}
\end{table*}

In Table \ref{tab:dd-cross-samples}, we show sample predictions from the DailyDialog++/cross task \cite{Sai2020ImprovingDE}. As DialoGPT has the best performance in the probe setup, on this task among the baselines, we choose it for error analysis. We randomly sampled 11 instances where the DialoGPT model made a mistake and observed the behavior of our DMI\_Base model on these samples. We see that our model correctly predicts for all 6 out of 6 negative samples and out of the 5 positive samples DMI\_Base predicts the label of 2 samples correctly (overall 8/11 correct predictions by our model). This shows that our model has a better understanding of the context and response inputs, which makes it robust against the adversarial negative samples. As can be seen in samples 2, 3, 5 and 6, the incorrect predictions by the DialoGPT model might have been caused by presence of common or similar meaning tokens
(\textit{cook, food; million; long; employee}) between context and response. This means that DialoGPT often relies on weak token-based cues for prediction.

For error analysis on the Empathetic-Intent (E-Intent) task \cite{welivita2020taxonomy}, we chose the ConveRT model as the baseline to compare against predictions from our DMI\_Base model. First, we randomly select 10 samples from the test set of the E-Intent task where the baseline ConveRT model makes a mistake. Then the predictions from the DMI\_Base model are observed
on these 10 samples. The input utterances, true labels and the predictions made by the model are shown in Table \ref{tab:eintent-samples}. Out of these 10 samples, DMI\_Base is able to predict the labels for 6 instances correctly. We notice that though sample inputs often contain more than one emotion, the one denoted by the gold label is generally the primary one. Our model is able to capture this emotion correctly more often than the baseline, with such mixed-emotion samples. 

\begin{table*}[]
\small
\centering
\begin{tabular}{@{}rL{7.8cm}rrr@{}}
\toprule
\textbf{ID} &\textbf{Input Utterance} & \textbf{Gold Label} & \textbf{ConveRT} & \textbf{DMI\_Base} \\ \midrule
\multicolumn{5}{c}{\textit{DMI\_Base predicts correctly}} \\ \midrule
1 & i feel very thankful for everything that i have, i live a really good life in my liking & grateful & content & grateful \\ \midrule
2 & I'm training a new girl at work. She is doing so good for her first week! & proud & confident & proud \\ \midrule
3 & It broke my heart today when I went to the grocery store and found out that they were out of Dean's French Onion Dip. & disappointed & devastated & disappointed \\ \midrule
4 & My wife's birthday is coming up. I got her a gift and the party planned out way ahead of time this year. & prepared & surprised & prepared \\ \midrule
5 & My friend helped me to pack & grateful & trusting & grateful \\ \midrule
6 & For two years now I've been walking with help of a walker, following a botched hip operation. Recently, at a physical therapy session, I was able to walk with a cane the length of the treatment room. I felt quite good about myself! & proud & caring & proud \\ \midrule
\multicolumn{5}{c}{\textit{DMI\_Base predicts wrongly}} \\ \midrule
7 & I was trying to plan my wedding by getting a caterer, and they kept blowing us off over and over again. & furious & disappointed & disappointed \\ \midrule
8 & Being a successful single mothr. & proud & content & content \\ \midrule
9 & We were over at our friend's house for a dinner and I was in the kitchen helping her cook. I had melted butter in a baking dish to make dessert, and I poured cold milk into it like the recipe said to do. It ended up cracking the dish. I felt bad. I offered to buy her a new one. & guilty & caring & ashamed \\ \midrule
10 & One time I had done really well in a class. I fully expected to get an A in it & anticipating & confident & disappointed \\ \bottomrule
\end{tabular}
\caption{Example Predictions on the Empathetic-Intent (E-Intent) task by ConveRT and our DMI\_Base model.}
\label{tab:eintent-samples}
\end{table*}

\begin{figure*}
    \centering
    \includegraphics[width=\textwidth]{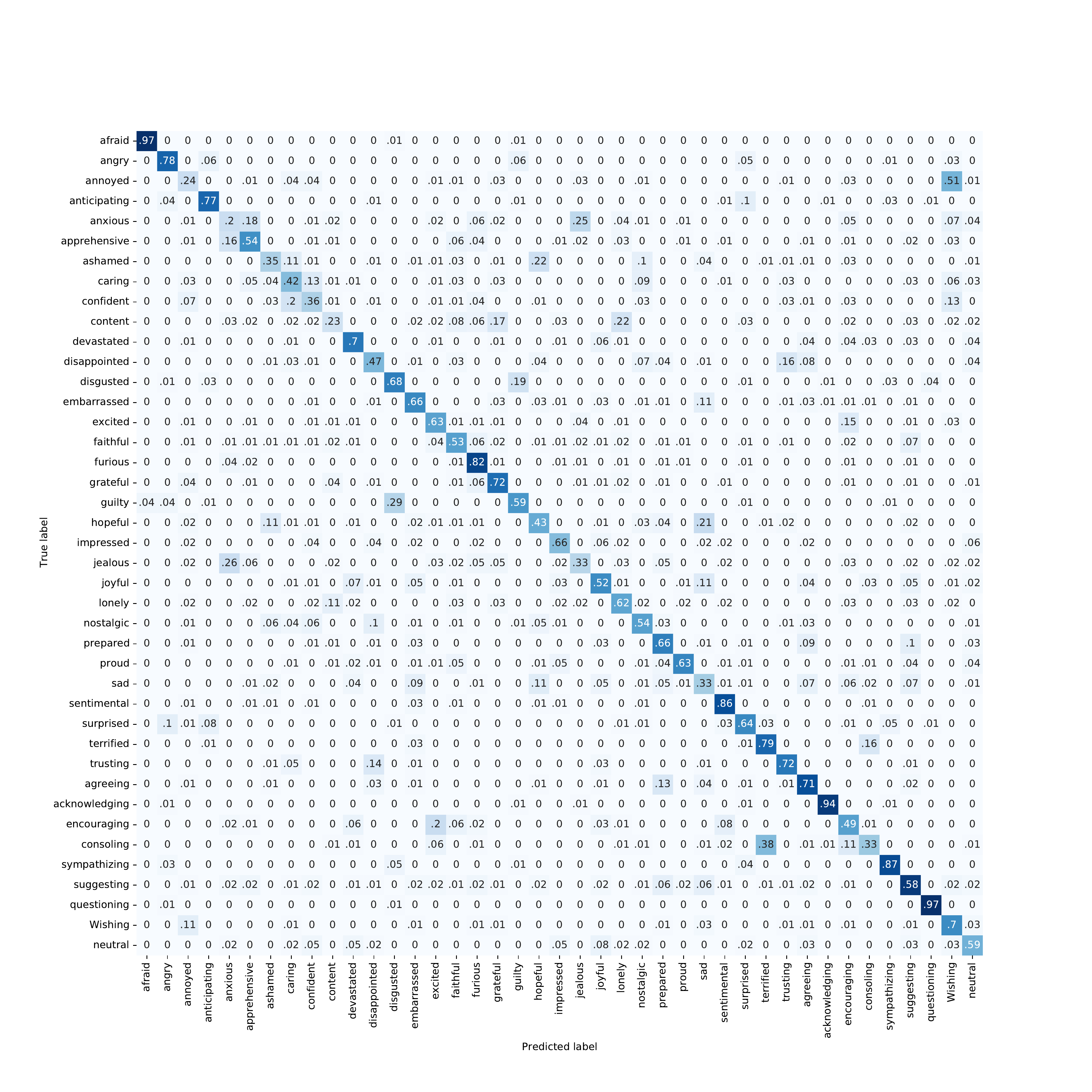}
    \caption{Confusion Matrix for our DMI\_Base model for the Empathetic-Intent task.}
    \label{fig:cm_eintent}
\end{figure*}

Fig.~\ref{fig:cm_eintent} shows the confusion matrix for our DMI\_Base model for the Empathetic-Intent task. The accuracy is highest for afraid, acknowledging and questioning classes (each above 95\%). Some of the most confusing pairs of classes are (annoyed, wishing), (anxious, apprehensive), (caring, confident), (content, grateful), (content, lonely). 

\end{document}